\documentclass[sigconf]{acmart}
\renewcommand\footnotetextcopyrightpermission[1]{} 
\usepackage{verbatim}
\usepackage{algorithm,setspace}
\usepackage[noend]{algpseudocode}
\usepackage{amsthm}
\usepackage{makecell}
\usepackage{subfigure}
\usepackage{graphicx}
\AtBeginDocument{%
  \providecommand\BibTeX{{%
    \normalfont B\kern-0.5em{\scshape i\kern-0.25em b}\kern-0.8em\TeX}}}

\begin{document}

\title{Multi-VFL: A Vertical Federated Learning System for Multiple Data and Label Owners}


\author{Vaikkunth Mugunthan}
\email{vaik@mit.edu}
\affiliation{%
  \institution{Massachusetts Institute of Technology}
  \city{Cambridge}
  \state{MA}
  \country{USA}
}

\author{Pawan Goyal}
\email{pawan14@mit.edu}
\affiliation{%
  \institution{Massachusetts Institute of Technology}
  \city{Cambridge}
  \state{MA}
  \country{USA}
}

\author{Lalana Kagal}
\email{lkagal@mit.edu}
\affiliation{%
  \institution{Massachusetts Institute of Technology}
  \city{Cambridge}
  \state{MA}
  \country{USA}
}

\renewcommand{\shortauthors}{Trovato and Tobin, et al.}

\begin{abstract}

Vertical Federated Learning (VFL) refers to the collaborative training of a model on a dataset where the features of the dataset are split among multiple data owners, while label information is owned by a single label owner. In this paper, we propose a novel method, Multi Vertical Federated Learning (Multi-VFL), to train VFL models when there are multiple data and label owners. Our approach is the first to consider the setting where $D$-data owners (across which features are distributed) and $K$-label owners (across which labels are distributed) exist. This proposed configuration allows different entities to train and learn optimal models without having to share their data. Our framework makes use of split learning and adaptive federated optimizers to solve this problem. For empirical evaluation, we run experiments on the MNIST and  FashionMNIST datasets. Our results show that using adaptive optimizers for model aggregation fastens convergence and improves accuracy.

\end{abstract}


\maketitle
\pagestyle{plain}

\section{Introduction}

Deep learning models benefit from large training datasets. However, for high-quality models, training needs to happen on large diverse datasets which usually are spread across multiple organizations. Deep learning solutions are extremely useful for for several domains including healthcare organizations and financial companies. While dealing with sensitive data (e.g.) patient information, credit card details, etc., government regulations such as GDPR \cite{voigt2017eu} and CCPA \cite{pardau2018california} restrict how entities can share data with other organizations. To solve this problem, Federated Learning (FL) was proposed. FL \cite{konevcny2016federated, mcmahan2017communication} is a collaborative learning paradigm that allows distributed clients to train machine learning models without having to share their sensitive data.

There are two main categories of FL. In horizontal FL, clients have the same set of features for a different set of clients, whereas, vertical FL handles the case when multiple clients possess the data of the same set of individuals, but each client has a unique set of features. There has been a lot of research in horizontal FL, however, there are few approaches for vertical FL \cite{liu2019communication,nock2018entity}. Split learning is the most commonly used approach to address vertical FL \cite{vepakomma2018split, ceballos2020splitnn,angelou2020asymmetric}. Split learning is a technique to collaboratively train a deep learning network that is split across different clients.  

Usually, vertical FL is used to solve the problem when dataset features are distributed across multiple data owners and there exists a single label owner. However, in reality, there might be multiple label owners. In this paper, we consider the setting where $D$ data owners (across which dataset features are distributed) and $K$ label owners (across which dataset labels are distributed) exist. For example, in the case of healthcare, the $D$ data owners may correspond to different specialist hospitals (heart disease hospital, lung disease hospital, cancer hospital, etc.) that have different data for the same set of patients and $K$ label owners correspond to COVID-testing centers that have the COVID results corresponding to these patients. 

To solve this problem, we make use of split learning and adaptive optimizer-based horizontal FL algorithms. Split learning solves the problem when $D$ data owners and $1$ label owner exists. However, when we have to aggregate the results from $K$ label owners, we need an aggregation mechanism. The most widely used FL algorithm for aggregation is FedAvg \cite{mcmahan2017communication}, where the weights of the models obtained from different label owners are just averaged. FedAvg is successful when label owners have IID (independent and identically distributed) data. However, in the real-world, label owners have non-IID data leading to client drift \cite{hsu2019measuring},  which is what happens when gradients computed by different label owners are skewed causing local models to move away from globally optimal models. This substantially affects the performance of FedAvg, especially in scenarios where label owners have a high degree of variance in their data distributions. Under such scenarios, adaptive learning rates and momentum are beneficial as they incorporate knowledge from prior iterations. When participating label owners have sparse data distributions and possess a limited subset of labels, SGD with server momentum improves the FL performance \cite{hsu2019measuring}. \cite{reddi2020adaptive} proposed federated versions of adaptive optimizers and showed that these optimizers improved the performance of FL under non-iid settings. To improve the performance of Adam in local settings, \cite{chen2019demon} applied a decaying momentum (Demon) rule to Adam. We extend this approach to the federated setting.

An entity resolution protocol such as Private Set Intersection (PSI) \cite{angelou2020asymmetric} is used to identify intersecting sets of individuals across different data owners and label owners by comparing the encrypted versions of the sets.

\section{Related Work}

Vertical FL is a distributed learning paradigm on vertically partitioned datasets. Conventional approaches for vertical FL make use of Multi-Party Computation and Homomorphic Encryption \cite{hardy2017private}. However, these approaches suffer from performance and communication problems. To address these issues, split learning was proposed. The primary benefit of using split learning is that participants don't have to share their sensitive data with each other while learning the shared model. 

\cite{ceballos2020splitnn} proposed a VFL technique using split neural networks for $D$ data owners and $1$ label owner. In their approach, each data owner trains a partial network up to a $\textit{cut}$ layer and sends the activations (outputs) of the $\textit{cut}$ layer to the label owner. The label owner concatenates the activations coming from different data owners and completes the forward pass. As the label owner has the labels corresponding to the individuals on the datasets possessed by the data owners, it carries out its backpropagation up to the $\textit{cut}$ layer. The gradients at the cut layer are split and passed to the corresponding data owners. This process repeats for a fixed number of rounds or till convergence is achieved. 

SplitFed, an approach that combined FL and split learning, was proposed by \cite{thapa2020splitfed}. Their framework was proposed for $D$ data owners and $1$ label owner as well. In their framework, a 3rd party server aggregates the models of different data owners using Federated Averaging (FedAvg). In addition, data owners update their models sequentially, hence increasing the total time per federated round. Also, they expect data owners to have the same model architecture which is a hard constraint on resource-deficit data owners.

Both papers consider $D$ data owners and $1$ label owner in their setting. In this paper, we propose a solution to train vertical FL models in which $D$ data owners and $K$ label owners exist.

\section{Framework Description}

The primary goal of our framework is to learn an optimized vertical FL model when $D$ data owners and $K$ label owners exist.  

In our framework, each data owner has a unique model architecture and all label owners have the same model architecture. Data owners perform forward propagation on their respective partial neural networks up to the $\textit{cut}$ layer and send their activations to label owners. Label owners concatenate the activations coming from different data owners and complete their forward pass. Label owners compute the loss, perform back-propagation, and send the corresponding gradients to data owners using which they complete their back-propagation. Label owners send their weights to the aggregation server which aggregates their weights using one of the following techniques: FedAvg \cite{mcmahan2017communication}, FedAdam \cite{reddi2020adaptive}, FedYogi \cite{reddi2020adaptive}, etc. We also extend FedAdam to  FedDemonAdam, a federated learning algorithm in which ''Demon\cite{chen2019demon} to Adam (DemonAdam)" is used for the server updates. The aggregation server returns the updated weights to the label owners which they use in the next round. The variables used in the Multi-VFL algorithm (Algorithm \ref{alg:vfl}) are provided in Table \ref{tab:algoparameters}.

\begin{figure*}
  \includegraphics[width=\textwidth]{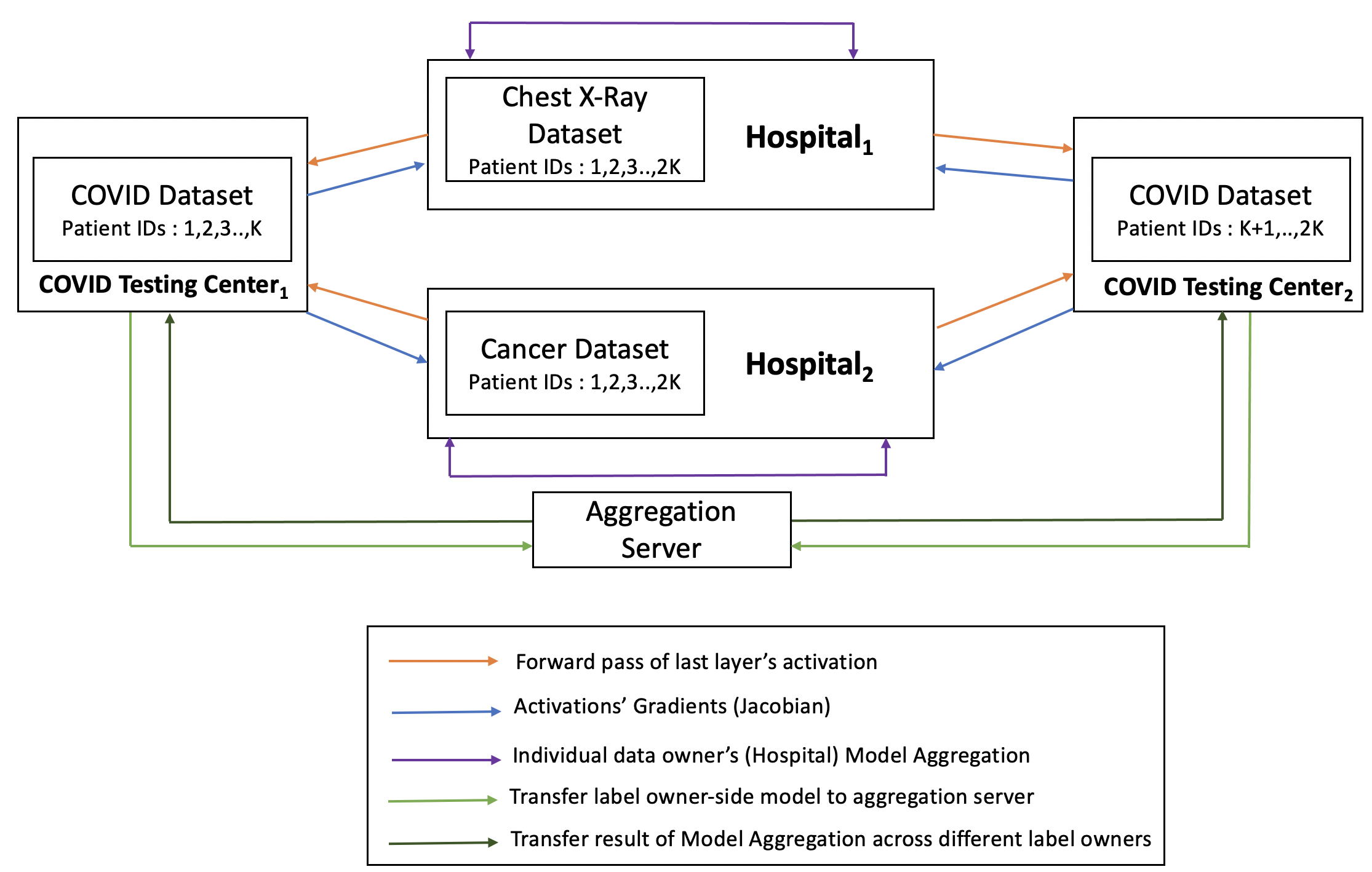}
  \caption{Real-Life Use Case for Multi-VFL}
  \label{fig:vfl}
\end{figure*}

An example real-life use case is depicted in Figure \ref{fig:vfl}. Let us consider two hospitals (data owners) and two COVID testing centers (label owners). Our goal is to develop a model that would be able to predict how likely a patient is to get COVID based on the information available from cancer and lung disease hospitals. Algorithm \ref{alg:vfl} provides an optimized solution for this scenario.

\begin{table}
\centering
 \begin{tabular}{c c} 
 \toprule
 \textbf{Variable}  & \textbf{Description}  \\ [0ex]
 \midrule 
 $r$ & Federated Learning Round \\ 
 $d, D$ & $D$ data owners, each indexed by $d \in {1,..,D}$ \\
 $k, K$ & $K$ label owners, each indexed by $k \in {1,..,K}$\\
 $\textbf{W}_{k}^{S,r}$ & Label Owner-side model of label owner $k$ at round $r$ \\
  $\textbf{W}_{d,k}^{C,r}$ & \makecell{ Owner-side model of data owner $d$ with \\ label owner $k$ at round $r$} \\
   $\textbf{W}_{d,k}^{C,r}$ & \makecell{ Owner-side model of data owner $d$ with \\ label owner $k$ at round $r$} \\
   $\textbf{A}_{d,k}^{S,r}$ & \makecell{ Activations from data owner $d$ to \\ label owner $k$ at round $r$} \\
  $\textbf{Y}_{d}$ & \makecell{True labels held by label owner corresponding to \\ individuals in Data owner $d$} \\
 \bottomrule
\end{tabular}
\caption{Variables used in our algorithm.}\label{tab:algoparameters}
\end{table}

\begin{algorithm}
\setstretch{1.3}
\caption{Multi-VFL with \textcolor{purple} {DemonAdam} / \textcolor{blue} {Adam} Server Aggregation.  }\label{alg:vfl}

\begin{flushleft}
\textbf{Initialization: } Initialize $\textbf{W}_{k}^{S,r}, \textbf{W}_{d,k}^{C,r} $ using Gaussian/Xavier initializer, Momentum parameters $\beta_1, \beta_2 \in [0,1)$, Server learning rate $\eta_s$, Stability constant $\mathcal{s}$.
\end{flushleft}
\begin{algorithmic}[1]

    \For{global round $r = 1,...R$}
        
        \For{each label owner $k \in K$ \textbf{in parallel}} 
            \State $\textbf{A}_{d,k}^r \leftarrow$ \textcolor{red}{$DataOwnerUpdate(d,r)$} $, \forall d \in D $ 
            \State $\textbf{A}_{k}^r \leftarrow $ Concatenate $\textbf{A}_{d,k}^r, \forall d \in D$  
            \State Forward propagation with $\textbf{A}_{k}^r$ on $ \textbf{W}_{k}^{S,r}$ and compute $ \hspace*{3em} \textbf{Y'}_{d}$
            \State Calculate loss with $ \textbf{Y'}_{d}$ and $ \textbf{Y}_{d}$
            \State Back-Propagation : $ \textbf{W}_{k}^{S,r+1} \leftarrow  \textbf{W}_{k}^{S,r} - \eta_k \nabla l(\textbf{W}_{k}^{S,r};\textbf{A}_{k}^r)$
            \State Send corresponding gradients to data owners by \hspace*{2.9em} invoking \textcolor{red}{$DataOwnerBackProp$(d$\textbf{A}_{d,k}^r$)},  $\forall d \in D$ 
            \State $\textbf{t}_{k}^{S,r+1}=\textbf{W}_{k}^{S,r+1}-\textbf{W}_{k}^{S,r}$
            \State $\textbf{W}_{k}^{S,r} \leftarrow$ \textcolor{red}{$ServerAggregation$ ($\textbf{t}_{k}^{S,r+1},opt$)}
            
        \EndFor
        
        \For{each data owner $d \in D$ \textbf{in parallel}} 
        
            \State $\textbf{W}_{d}^{C,r+1} \leftarrow \frac{\sum_{k=1}^{K} \textbf{W}_{d,k}^{C,r}}{K}$
            \State $\textbf{W}_{d,k}^{C,r+1} \leftarrow \textbf{W}_{d}^{C,r+1}, \forall k \in K $
        \EndFor

    \EndFor
        
\end{algorithmic}

\begin{flushleft}

\textcolor{red}{$DataOwnerUpdate(d,r):$}\\
\begin{algorithmic}[1]
\State $\textbf{A}_{d}^{r} = \phi$
    \For{local epoch $e=1,..E$}
        \For{batch $b \in B$}
            \State Perform Forward propagation on $\textbf{W}_{d}^{C,r}$
            \State Concatenate final layer activations to $\textbf{A}_{d}^{r}$
        \EndFor
    \EndFor
\State Return $\textbf{A}_{d}^{r}$    
\end{algorithmic}

\end{flushleft}

\begin{flushleft}

\textcolor{red}{$DataOwnerBackProp$(d$\textbf{A}_{d,k}^r$)}\\

\begin{algorithmic}[1]

        \For{batch $b \in B$}
            \State Perform back propagation with d$\textbf{A}_{d,k}^r$
            and update $\textbf{W}_{d}^{C,r}$
        \EndFor

\end{algorithmic}

\end{flushleft}

\begin{flushleft}
\textcolor{red}{ $ServerAggregation(t_{k}^{S,r},opt) $ } 

\begin{algorithmic}[1]
\If{\textcolor{purple} {$opt=demon$}}
            
            \State \textcolor{purple}{$\beta_1 \gets \beta_1 \frac{(1 - \frac{r}{R})}{(1 - \beta_1) + \beta_1 ( 1 - \frac{r}{R})}$}
            \State \textcolor{purple}{$m \gets \beta_1 m + \textbf{t}_{k}^{S,r}$ }
        \Else
        \State \textcolor{blue}{$m \gets \beta_1 m + (1-\beta_1) \textbf{t}_{k}^{S,r}$}
        \EndIf
        \State $v \gets \beta_2 v + (1-\beta_2) m^2$ 
        
        \State $\textbf{W}^A_{r+1}=\textbf{W}^A_r + \frac{\sqrt{1-\beta_2^r}}{1-\beta_1^r}\eta_s\frac{m}{\sqrt{v} + \mathcal{s}}$ 
        \State Return $\textbf{W}^A_{r+1}$

\end{algorithmic}
\end{flushleft}

\end{algorithm}

\section{Experiments}
To verify the validity of our proposed model, we ran experiments on two major datasets: MNIST \cite{lecun1998mnist} and FashionMNIST \cite{xiao2017fashion}. Both datasets have 60,000 training data points and 10,000 testing data points with each image as $28\times 28$. For our experiments, we consider  5 label owners and 4 data owners. The datasets (excluding labels) are vertically partitioned among the data owners, with each of them receiving images of dimension $(\frac{28}{4} = 7) \times 28$. The dataset labels are with the label owners. Each data owner has one convolutional layer with input and output channels as 1 and 32 respectively, and kernel size as 3. The output from the data owners are aggregated to form a tensor with 32 channels which are then passed to the label owner's model which has one convolutional and two linear layers. All the experiments are run for 500 epochs with a batch size of 64. The local learning rate was set to 0.001. We study four major federated learning aggregation algorithms: FedAvg (Base Case), FedAdam, FedYogi, and FedDemonAdam. The values of the hyperparameter for these algorithms are summarized in Table \ref{Optimizer's Hyperparameters}.

\begin{table}[h]
\centering
\begin{tabular}{|c|cccc|}
\hline
         Optimizer    & beta 1 & beta 2 & lr   & tau  \\ \hline
\makecell{ FedAdam, FedYogi \\ FedDemonAdam }      & 0.9    & 0.99   & 1e-3 & 1e-3 \\
 \hline
\end{tabular}
\caption{Optimizer's Hyperparameters}
\label{Optimizer's Hyperparameters}
\end{table}

\vspace{-1.9em}
We also allow different label owners to have different labels, thus mimicking the non-iid nature of the real world. Each label owner is given 5000 data points to train upon. However, some label owners are given an iid dataset, meaning all labels from 0 to 9 are given, while others are given only two labels each. We primarily considered 4 different ways to distribute labels and they are formulated in Table \ref{non-iid setup}.

\begin{table}[h]
\centering
\begin{tabular}{|c|ccccc|}
\hline
     Scenario & \multicolumn{1}{c}{\makecell{ Label \\ Owner 1 }} & \multicolumn{1}{c}{\makecell{ Label \\ Owner 2 }} & \multicolumn{1}{c}{\makecell{ Label \\ Owner 3 }} & \multicolumn{1}{c}{\makecell{ Label \\ Owner 4 }} & \makecell{ Label \\ Owner 5 } \\ \hline
1niid & 0-9                           & 0-9                           & 0-9                           & 0-9                           & \{0,1\}  \\
2niid & 0-9                           & 0-9                           & 0-9                           & \{0,1\}                       & \{2,3\}  \\
3niid & 0-9                           & 0-9                           & \{0,1\}                       & \{2,3\}                       & \{4,5\}  \\
4niid & 0-9                           & \{0,1\}                       & \{2,3\}                       & \{4,5\}                       & \{6,7\}  \\ \hline
\end{tabular}
\caption{non-iid setup}
\label{non-iid setup}
\end{table}
\vspace{-1em}

\begin{figure*}
  \centering
  \subfigure[MNIST]{\includegraphics[scale=0.49]{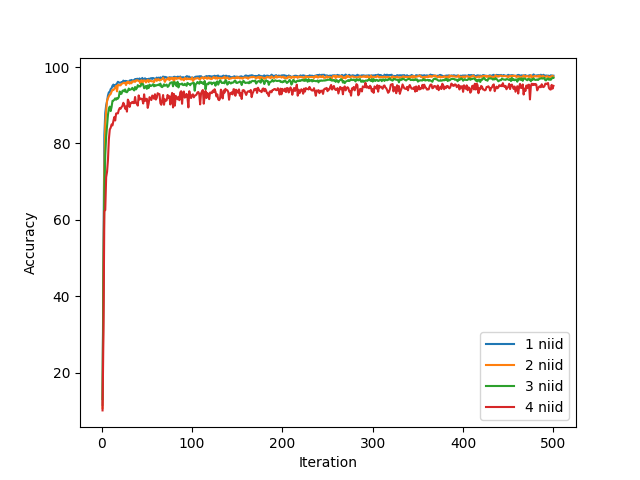}}\quad
  \subfigure[FashionMNIST]{\includegraphics[scale=0.49]{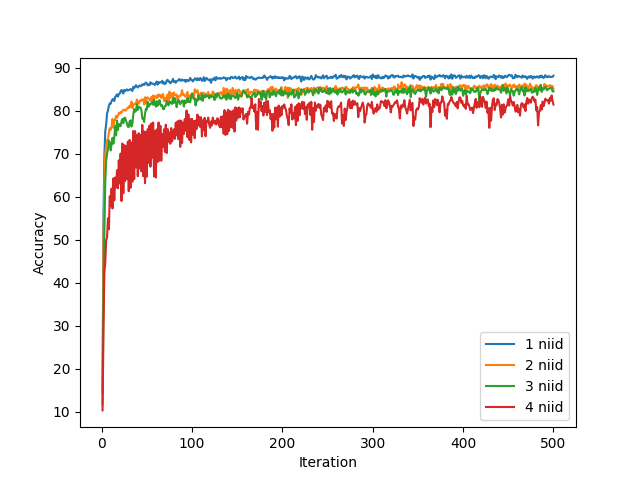}}
  \vspace{-10pt}
  \caption{Variation of accuracy with number of non-iid label owners}
  \label{sgd_niid}
\end{figure*}

\begin{figure*}
  \centering
  \subfigure[MNIST]{\includegraphics[scale=0.49]{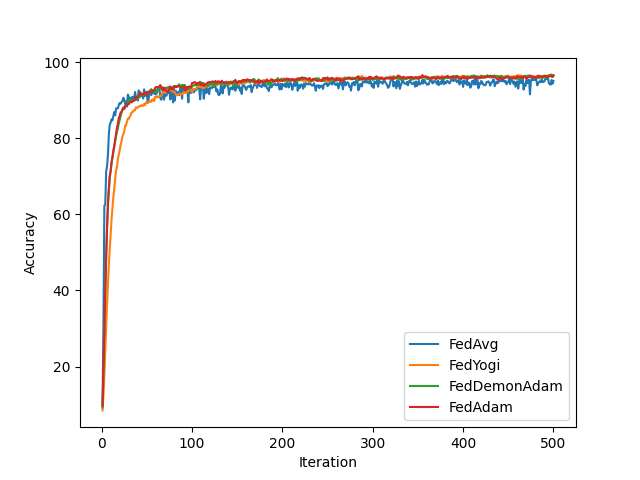}}\quad
  \subfigure[FashionMNIST]{\includegraphics[scale=0.49]{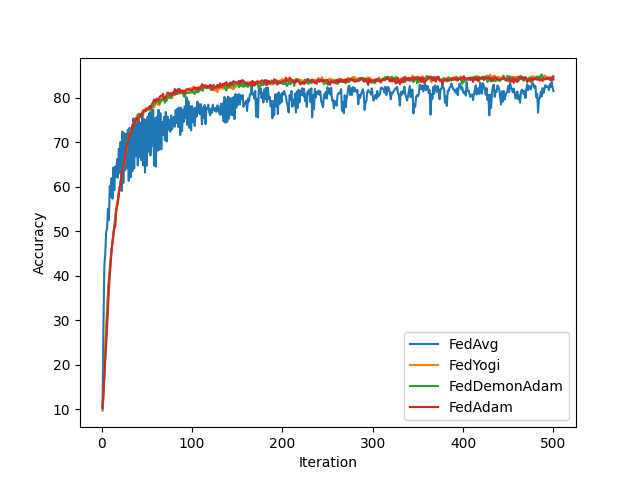}}
   \vspace{-10pt}
  \caption{Variation of accuracy for different optimizers}
  \label{4niid optim}
\end{figure*}

\subsection{Variation with number of non-IID servers}
We first study the variation of accuracy with increase in the number of non-iid label owners. Primarily, we study the 1niid, 2niid, 3niid and 4niid scenarios and use the FedAvg algorithm to solely demonstrate the effect of non-iid-ness without any optimizers. In FedAvg, the aggregation server averages the weights collected from different label owners and sends them back. The label owners then update their weights. The results for MNIST and FashionMNIST are plotted in Fig \ref{sgd_niid}.

As seen from Fig \ref{sgd_niid}, the accuracy drops with and increase in the number of non-iid label owners. In addition, for the 4niid case, FedAvg fails to converge even after 500 iterations.

\subsection{Variation with optimizers}

In this experiment, we study the variation of accuracy for the 4niid scenario with different choices of optimizers. The results for MNIST and FashionMNIST are plotted in Fig \ref{4niid optim}. As clear from the figure, adaptive optimizers perform better and converge faster than the FedAvg algorithm, with greater effect in the FashionMNIST dataset where they improve the accuracy by 2-3\%. Also, we see that FedAvg fails to converge for the FashionMNIST dataset.

\section{Conclusion and Future Work}

To the best of our knowledge, Multi-VFL is the first solution to address vertical federated learning when $D$ data owners and $K$ label owners exist. In addition, we ran experiments on the MNIST and FashionMNIST datasets for different non-IID label distribution scenarios and demonstrated the importance of the adaptive optimizer based aggregation.  We plan to integrate differential privacy \cite{dwork2008differential} into our framework to thwart potential model inversion attacks and plan to conduct a detailed analysis on the privacy-accuracy trade-off.

\newpage
\bibliographystyle{ACM-Reference-Format}
\bibliography{sample-base}

\end{document}